\documentclass{svproc}
\usepackage{url}

\usepackage{dblfloatfix} 
\usepackage{subcaption}
\usepackage{multirow}
\usepackage{tabularx,booktabs,hhline} 
\usepackage{algcompatible}
\usepackage{caption}
\usepackage{hyperref}
\usepackage[linesnumbered,ruled,vlined]{algorithm2e}
\SetKwInput{KwInput}{Input}                
\SetKwInput{KwOutput}{Output}              
\usepackage{amsmath}
\usepackage[markup=underlined]{changes}

\begin{document}
\mainmatter              
\title{Enhancing Explainability in Mobility Data Science through a combination of methods}
\titlerunning{ Pre-print Explainability in Mobility Data}  

\author{Georgios Makridis\inst{1} \and Vasileios Koukos \inst{1}
Georgios Fatouros \and Dimosthenis Kyriazis }
\authorrunning{Georgios Makridis et al.} 
%
%
\institute{Department of Digital Systems, University of Piraeus, Pireas Karaoli ke Dimitriou 80 185 34, Greece,\\
\email{gmakridis@unipi.gr}}

\maketitle              

\begin{abstract}
In the domain of Mobility Data Science, the intricate task of interpreting models trained on trajectory data, and elucidating the spatio-temporal movement of entities, has persistently posed significant challenges. Conventional XAI techniques, although brimming with potential, frequently overlook the distinct structure and nuances inherent within trajectory data. Observing this deficiency, we introduced a comprehensive framework that harmonizes pivotal XAI techniques: LIME (Local Interpretable Model-agnostic Explanations), SHAP (SHapley Additive exPlanations), Saliency maps, attention mechanisms, direct trajectory visualization, and Permutation Feature Importance (PFI). Unlike conventional strategies that deploy these methods singularly, our unified approach capitalizes on the collective efficacy of these techniques, yielding deeper and more granular insights for models reliant on trajectory data. In crafting this synthesis, we effectively address the multifaceted essence of trajectories, achieving not only amplified interpretability but also a nuanced, contextually rich comprehension of model decisions.

To validate and enhance our framework, we undertook a survey to gauge preferences and reception among various user demographics. Our findings underscored a dichotomy: professionals with academic orientations, particularly those in roles like Data Scientist, IT Expert, and ML Engineer, showcased a profound, technical understanding and often exhibited a predilection for amalgamated methods for interpretability. Conversely, end-users or individuals less acquainted with AI and Data Science showcased simpler inclinations, such as bar plots indicating timestep significance or visual depictions pinpointing pivotal segments of a vessel's trajectory. Notably, the survey highlighted a unanimous appreciation for juxtaposing predicted versus actual trajectories as a direct benchmark of model performance. Furthermore, visualizations emphasizing critical past vessel positions were hailed, with technical roles finding them especially enlightening and end-users perceiving them as intuitive and enlightening. Our tabled results provide a more detailed breakdown of XAI usability preferences in Vessel Route Forecasting (VRF), further enriching our contributions to the field. 

\end{abstract}
%

\section{Introduction}
\label{intro}
\subsection{Background}
Trajectory data, representing the movement of entities over time and space, has become indispensable in a world that's increasingly interconnected and monitored. From tracking the migratory patterns of endangered species to monitoring ships across oceans, trajectories provide a dynamic representation of how entities interact with the environment. GeoXAI, as an emerging field, seeks to bridge the chasm between advanced machine learning techniques and geospatial data interpretation.
Yet, with the richness of trajectory data comes complexity. Traditional machine learning techniques, though effective, often work as black boxes, obfuscating the logic behind their predictions \cite{jalali2023towards}. In the context of geospatial analysis, such opacity is especially problematic, given the potential real-world implications of the decisions drawn from these models \cite{wiegreffe2019attention}.
While most eXplainable AI (XAI) research focuses on areas like computer vision, natural language processing, and sectors using structured data (like healthcare \cite{liu2023predicting} and finance \cite{kotios2022deep}) or even in cybersecurity tasks \cite{makridis2022xai}, there are limited studies that apply XAI techniques to GeoAI (termed as GeoXAI) \cite{hsu2023explainable}.
Moreover, the intricate nature of trajectory data, encompassing both spatial and temporal dimensions, poses unique challenges for model interpretability \cite{xing2023challenges}. There's an evident gap: the need for transparent, interpretable models tailored to the nuances of trajectory data.

\subsection{Challenges of GeoXAI for Trajectory Data Interpretation}

Understanding the complexity of integrating XAI methods into GeoAI, especially in the fields of Geographic Information System (GIS) science and Mobility Data Science (MDS), necessitates a deep grasp of both XAI techniques and the information embedded in spatio-temporal data. Trajectory data, a sub-domain of spatio-temporal information, carries its unique set of challenges when paired with XAI. Here are the key challenges as described in \cite{jalali2023towards}:

\begin{itemize}
    \item \textbf{Selection of Reference Data and Models:} Choosing suitable reference data and models is paramount.
    \item \textbf{Gradients as Explanations:} Utilizing gradients to pinpoint feature importance or influence can be inherently limited in capturing the multifaceted nature of spatial relationships and trajectories.
    \item \textbf{Geographic Scale Dilemmas:} Addressing scale is complex in geography. The challenge lies in determining the appropriate scale at which to interpret and explain trajectories, considering trajectories might look different when viewed at varying scales.
    \item \textbf{Integration of Topology and Geometry:} Current XAI tools, like LIME and SHAP, face challenges in incorporating the geometric and topological characteristics inherent in spatial data, which are essential for accurate trajectory interpretation.
    \item \textbf{Visualization of Geography in XAI:} Presenting geographical data within XAI outputs in a way that's intuitive yet detailed is a significant challenge. Traditional XAI visual methods might not suffice for the nuanced requirements of geospatial data.
    \item \textbf{Harnessing Geospatial Semantics and Ontologies:} Geospatial data is rich with semantics - the meaning associated with locations, paths, and patterns. Current XAI methods might not fully exploit this depth, missing out on leveraging geospatial ontologies to enhance explanations.
\end{itemize}

\subsection{Towards an Integrated GeoXAI Framework for Trajectory Forecasting Insights}

\par Recognizing these challenges, we set forth a clear objective: to design a GeoXAI framework that combines the power of cutting-edge machine-learning techniques with the specificity required for trajectory data interpretation. We aim to not only model trajectory data effectively but to do so transparently, ensuring that stakeholders, be they urban planners, conservationists, or maritime authorities, can get clear, actionable insights from these models. To address this, we propose a unified XAI approach that modifies well-established methods, including LIME, SHAP, Saliency maps, attention mechanisms, direct trajectory visualization, and Permutation Feature Importance, to suit trajectory data.

\section{GeoXAI and Trajectory Data: A Landscape Overview}

\subsection{GeoXAI applications: state-of-the-art}

GeoAI and XAI are just starting to work together, mostly to understand how well GeoAI is doing its job. For example, in \cite{luo2021glassboxing} the researchers looked at how to use AI for spotting aircraft but ran into issues like wrongly identifying planes because of nearby vehicles. This problem may be solved by the Maximum Mean Discrepancy- Critic algorithm (\cite{kim2016examples}), where high-quality explanation examples with various sizes of vehicles and aircraft are shown in relation to each other, although training examples are not easy to obtain. In \cite{behl2021twitter} XAI is used to check why some tweets about earthquakes were wrongly labeled. In another study,\cite{zhou2022identifying} AI and XAI tools [LIME \cite{ribeiro2016should} and Layer-Wise Relevance  Propagation (LRP)], were used to better place tweets on a map, but there's still work to be done to consider the full picture of a location.

There's a lot of activity in using AI to understand satellite pictures. For instance, \cite{abdollahi2021urban} and \cite{guo2021network}, used XAI to understand plants and land uses from images. SHAP \cite{lundberg2017unified} and LIME are the main tools everyone's using. For instance, in \cite{yang2021application} study showed how SHAP can help understand where truck accidents might happen, and in \cite{li2022extracting} they used it to guess where people might want a ride. However, the \cite{amiri2021peeking} study showed these tools don't always consider everything about a place, like its layout.

Lastly, many are using visuals like maps to make AI's decisions clearer. For instance, \cite{zhou2021salience} used color-coded maps. But, there's a push to make these visuals show actual places on Earth more accurately. In a more recent approach \cite{altieri2023explainable}  introduces a model-agnostic XAI method tailored for geo-distributed sensor network data, revealing influential factors through feature perturbations and offering explanations across features, timesteps, and locations, with demonstrated efficacy in real-world forecasting datasets

\subsection{The Unique Attributes of Trajectory Data in Geospatial Contexts}

Trajectory data is inherently dynamic, capturing the movement of objects or entities over space and time. This form of data, sourced from a myriad of technologies such as GPS, satellite tracking, and IoT devices, has spatial, temporal, and thematic dimensions \cite{laube2002analyzing}. Unlike static geospatial data, trajectories elucidate patterns of movement, interaction, and behavior \cite{andrienko2013visual}. Their multifaceted nature poses both challenges and opportunities for GeoXAI applications.

\subsection{Identified Gaps and the Need for an Advanced Framework}

Despite the advancements in GeoXAI, a noticeable gap remains in its application to trajectory data. Most existing methods, designed for static geospatial datasets, are not actually able to handle the complexity of trajectories. This inadequacy underscores the urgent need for an advanced framework that can integrate and combine the strengths of various methods of XAI with the complexity of trajectory data.

GeoXAI has consistently led efforts to enhance interpretability in geospatial data analytics, yet, when it comes to trajectory data, the adoption of these techniques is still in its infancy \cite{xing2023challenges}. To address the intricacies of geospatial trajectories, we propose a comprehensive approach that merges LIME, Saliency Maps, Permutation Feature Importance (PFI), Attention, and trajectory visualizations, specifically tailored for trajectory data. This combination allows users to leverage and merge insights from each method, optimizing their analytical outcomes. Herein, we dive deeper into our adaptations, showcasing the advantages and considerations of each combination.

\section{The Integrated GeoXAI Framework for Trajectory Interpretability}

\subsection{Dataset Used}

The dataset is derived from \cite{tritsarolis2022piraeus}, which contains vessel position information transmitted by vessels of different types and collected via the Automatic Identification System (AIS).

\subsection{Models Used}

A general methodology of Cross Industry Standard Process for Data Mining (CRISP-DM) was used to developed the data-driven approach.

By leveraging sequence models for trajectory data, some challenges arise. The aim is not only to capture the sequence intricacies but also to bring about model explainability. During the model-building phase, we used Bidirectional LSTMs. This approach effectively wraps around an LSTM layer, making it bidirectional and ensuring the availability of both forward and backward hidden states. The use of Bidirectional LSTMs is particularly beneficial for trajectory data. While a standard LSTM processes sequences from the past to the future, a Bidirectional LSTM processes the data in both directions. This ensures a more comprehensive representation, crucial for the intricate patterns in trajectory data.

Moreover, an integrated attention mechanism enhances the model's performance and explainability. It allows the model to dynamically prioritize different parts of the trajectory, offering clearer insights into which waypoints or segments the model finds most relevant during its predictions. This promotes model transparency.

To this end, the architecture of the utilized model can be summarized as :

\begin{itemize}
    \item Input layer: The input layer accepts sequence data with a shape determined by the number of time steps and features in the trajectory data. This ensures the model can process the entire sequence for each trajectory data point, capturing details like $(\Delta \text{lon}, \Delta \text{lat}, \Delta t_{\text{curr}}, \Delta t_{\text{next}})$.
    \item Bidirectional LSTM Layer: Directly following the input layer is the Bidirectional LSTM layer. This layer processes the input sequence in two directions:
    \begin{itemize}
    \item \textbf{Forward LSTM}: Processes the sequence from the start to the end, capturing forward temporal dependencies.
    \item \textbf{Backward LSTM}: Processes the sequence from the end to the start, capturing backward temporal dependencies.
    \end{itemize}
    \item Bahdanau Attention Mechanism: In our extended models, we have integrated an attention mechanism. This allows the model to weigh different parts of the trajectory data sequence differently, focusing on more relevant waypoints or segments during prediction. The Bahdanau attention \cite{bahdanau2014neural} was introduced to address the bottleneck performance of conventional encoder–decoder systems, resulting in significant improvements over the traditional technique. Since it computes linear combinations of encoder and decoder states, therefore it is also called an additive attention mechanism. The basic principle of the Bahdanau attention mechanism is to focus on particular input vectors of the input sequence on the basis of attention weights.
    \item Output layer: The concatenated hidden states from the forward and backward LSTMs serve as the input to the final layers of the network. A fully connected (Dense) layer, typically with a linear or appropriate activation function, produces the final output, such as predicted $(\Delta \text{lat}, \Delta \text{lon})$
\end{itemize}



\subsection{LIME for Spatial Trajectory Insight}

\begin{figure*}[h!]
\centering
\includegraphics[width=0.9\textwidth]{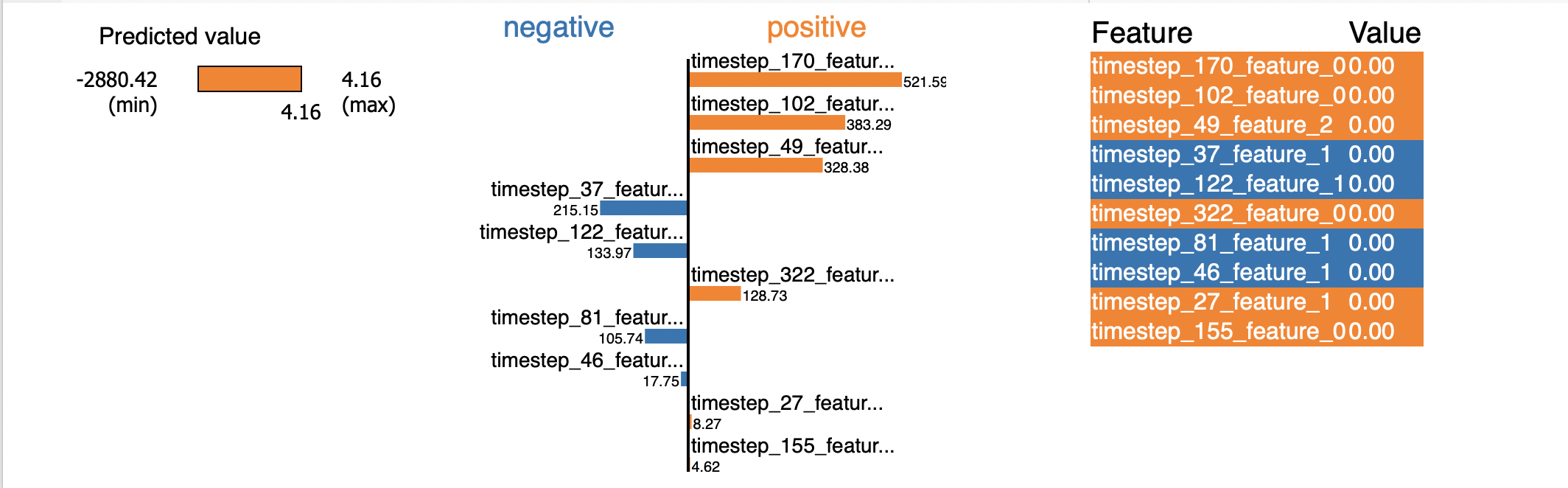}
\caption{ A result of the lime model applied to an unknown instance. If a timestep is shown in "blue" and has a large positive value, it means this feature significantly increased the model's prediction for that instance. A feature in "red/orange" would indicate that it decreased the prediction. The magnitude of the value will indicate the strength of the contribution. }
\label{fig:lime}
\end{figure*}

To apply LIME to the LSTM model, we face a challenge in that LIME expects input as 2D arrays (samples, features), while LSTM works with 3D inputs (samples, timesteps, features). This results from LIME being originally designed for classical models, which take tabular data as input. As a workaround, we created a wrapper function around the LSTM model's prediction function, which reshapes the 2D input from LIME to the 3D input that the LSTM model expects. As a result, the features that are consistently showing up in the top contributions can tell us which aspects of the data the model is focusing on. In a flattened time series, these are specific timesteps. If a feature is shown in "blue" and has a large positive value, it means this feature significantly increased the model's prediction for that instance. A feature in "red/orange" would indicate that it decreased the forecast. The magnitude of the value will indicate the strength of the contribution. In Fig. \ref{fig:lime} is depicted as one result of the lime model applied to an unknown instance.

The challenge lies in the fact that we have flattened the time series data, which might make it somewhat difficult to infer clear patterns. We'll have to correlate the timesteps and features back to your actual sequence.

\subsection{Saliency Maps for Spatial Trajectory Insights}

The notion of {Saliency maps} was first introduced by \cite{simonyan2013deep} Simonyan and Zisserman as a technique to enhance human cognition when dealing with complex image classification tasks in ML/AI systems. Saliency maps are designed to capture human attention in specific regions of the generated image, by identifying “special” features/pixels which are highlighted based on their underlying importance. To that end, human experts could have a better understanding of the classification output of a given ML/AI solution, thus making human-computer interaction a straightforward task.

To compute the Saliency Map in our task, the gradients of the output concerning the input should be depicted. So there should be a minor transformation to the code to work with an LSTM model. Specifically, this modification will return a 2D array of the same shape as the input data, with each value representing the "importance" of the corresponding timestep in the input for the model's output prediction. Higher values indicate timesteps that, when changed slightly, would result in a larger change in the output (and are therefore more "important" to the prediction).

\begin{figure*}[h!]
\centering
\includegraphics[width=0.8\textwidth]{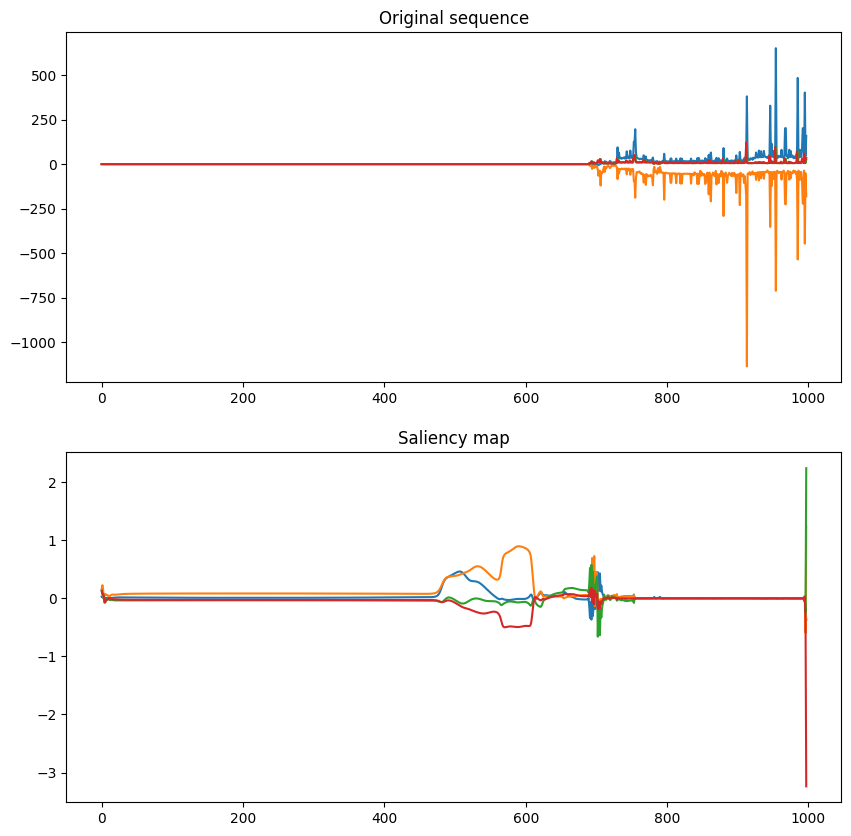}
\caption{The first plot  shows the original sequence. The second plot, the saliency map, shows the importance of each timestep in the sequence according to the model. }
\label{fig:saliency}
\end{figure*}

The first plot in \ref{fig:saliency} shows the original sequence. The second plot, the saliency map, shows the importance of each timestep in the sequence according to the model. Large absolute values in the saliency map correspond to timesteps in the sequence that the model paid a lot of attention to when making its prediction.

However, the interpretation of the saliency maps in the context of trajectory forecasting might not be straightforward. The saliency map will give the importance of each timestep, but due to the complex nature of sequence data and the LSTM model, it can be challenging to derive clear and easily understandable insights from these maps.

\subsection{Attention Mechanisms}

\begin{figure*}[h!]
\centering
\includegraphics[width=0.8\textwidth]{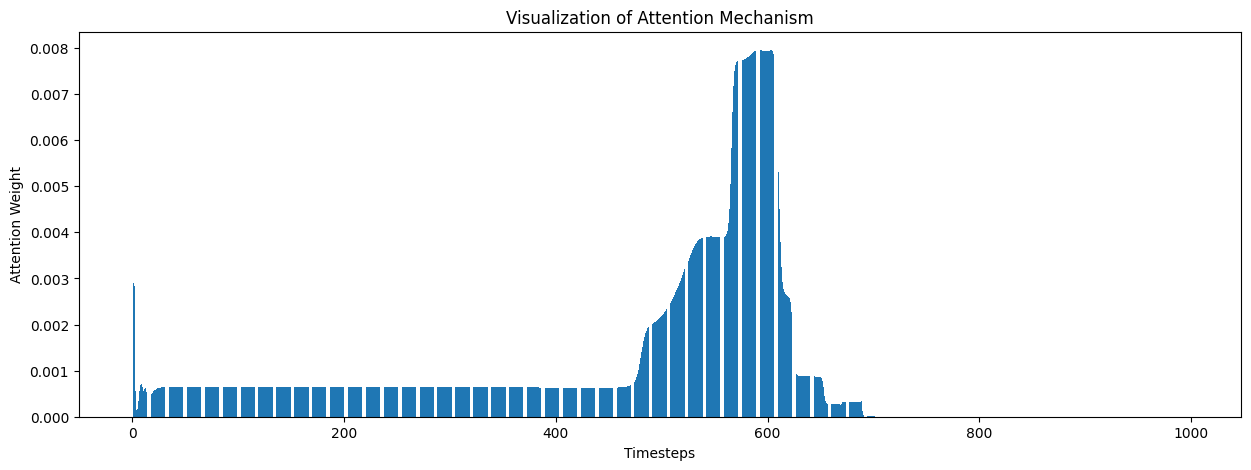}
\caption{Attention weights, a higher weight implies that the model found that particular timestep more important for its prediction. Of course, it will be common in a vessel position prediction, that the attention weights are higher for recent timesteps, this could suggest that the model has learned to pay more attention to the vessel's more recent positions when predicting its next position.
}
\label{fig:attention}
\end{figure*}

In a sequence-to-sequence prediction task like route forecasting, an attention mechanism helps the model focus on different parts of the input sequence when predicting each output time step. This is especially useful when the sequences are long, and different parts of the sequence might be more relevant for predicting different output time steps. In the attention weights plot, each bar represents a timestep in the input sequence, and the height of the bar indicates the attention weight that the model has assigned to that timestep. This plot (Fig. \ref{fig:attention} shows the attention weight assigned to each timestep in the sequence, which indicates how much the model "paid attention" to each timestep when making the predictions. A higher weight implies that the model found that particular timestep more important for its prediction. Of course, it will be common in a vessel position prediction, that the attention weights are higher for recent timesteps, this could suggest that the model has learned to pay more attention to the vessel's more recent positions when predicting its next position.

\subsection{Visualization Trajectories}

When visualizing trajectories, especially to assess the accuracy and reliability of predictive models, it's beneficial to overlay both the predicted and actual routes on a single plot. This provides immediate insights into the model's performance by highlighting areas of deviation and alignment. By using distinguishable line styles or colors for each route, users can swiftly gauge the fidelity of the prediction to the actual movement path. This approach not only offers a direct visual comparison but also aids in identifying specific segments or junctions where the model might consistently underperform.

\begin{figure*}[h!]
\centering
\includegraphics[width=0.6\textwidth]{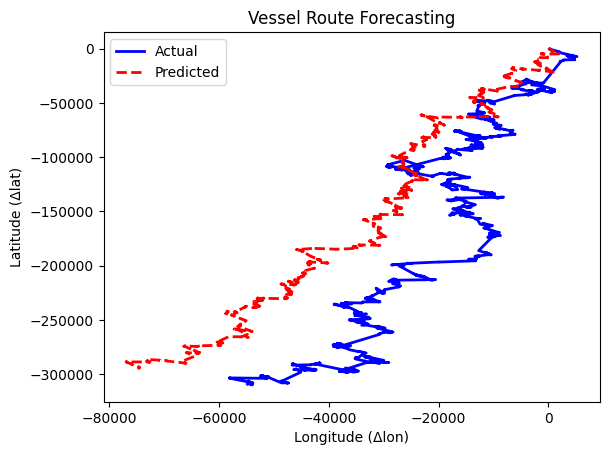}
\caption{ Predicted vs. Actual trajectory of a vessel}
\label{fig:trajectory}
\end{figure*}

An example of such a plot that contained the predicted and the actual trajectory of a vessel is depicted in \ref{fig:trajectory}.

\subsection{Permutation Feature Importance}

PFI is a technique used to determine the importance of individual features in a predictive model. By systematically shuffling values of a single feature and measuring the degradation in model performance, PFI measures the feature's significance. Features causing a notable decrease in model accuracy when permuted are deemed important, whereas features causing little to no change are considered less influential.

Given the Fig. \ref{fig:pfi} we can infer the following:

Latitude Importance: Since this has the highest score, it implies that changes in latitude are the most influential factor in predicting vessel movement. This could mean that vessels in your dataset are more frequently changing their north-south position, or that north-south movements have more significant consequences in terms of route changes.

Longitude Importance: The importance of longitude indicates that east-west movements are also essential in predicting the vessel route but are less critical than latitude changes. Depending on the regions and the type of routes that vessels are taking, this could be an expected behavior.

Current and Next Time Importance: The relatively lower importance of the time-related features might mean that the temporal aspects of the route (like how quickly a vessel is expected to move to the next point) are not as vital as the spatial coordinates. This might suggest that the physical location of the ship and the direction it's heading play a more dominant role in determining its future route. The insights drawn from these importance scores might help you understand the underlying patterns in vessel movements in the geographical areas you are analyzing. For example, certain sea routes might be more constrained in their east-west movements, reflecting natural barriers or common shipping lanes.

\begin{figure*}[h!]
\centering
\includegraphics[width=0.8\textwidth]{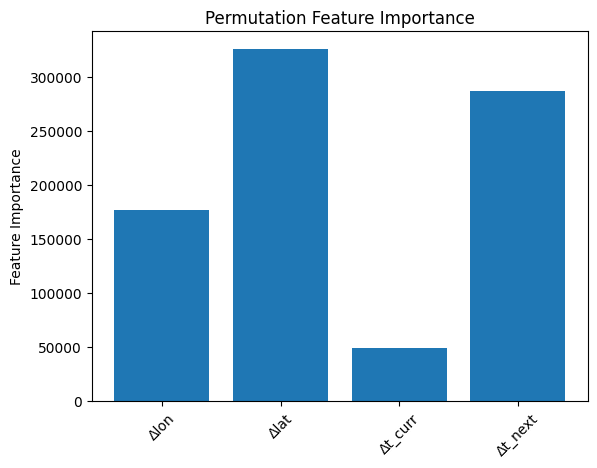}
\caption{ Bar Plot of Permutation Feature Importance of the 4 features of the trajectory dataset }
\label{fig:pfi}
\end{figure*}

\section{Proposed Approach}

As highlighted in the prior section, LSTMs remain at the forefront of vessel route forecasting models due to their exceptional prowess in managing time-sequenced data. However, with their superior performance comes a notable complexity that often cloaks their inner workings, rendering them as "black boxes". Given the profound implications of vessel route forecasting — from logistics and environmental facets to maritime safety — it's indispensable to unpack these models' decision-making intricacies.

To chart a course through this interpretability challenge, we embarked on an integrative methodological voyage. We combined both model-specific techniques like attention mechanisms with model-agnostic ones, notably LIME and SHAP. The attention mechanism anchors our approach, elucidating the significance of each timestep within sequences. Concurrently, LIME offers a panoramic, model-agnostic viewpoint, spotlighting the importance of distinct features in individual predictions. This juxtaposition enabled us to graphically represent maritime trajectories, accentuating critical timesteps using a vibrant color gradient, thereby offering a visually intuitive understanding.

Yet, our pursuit of interpretability didn't halt there. We further enriched our approach with SHAP, a cohesive framework that not only quantifies feature importance but also unravels inter-feature dependencies. This method deconstructs predictions into an aggregation of feature contributions, thus bestowing clarity on pivotal features and their intricate interplays in model decisions. Given our input's high-dimensional nature, we harnessed a bespoke Deep SHAP variant, aptly fashioned for deep learning architectures like our bidirectional LSTM. A curated subset of training data acted as our backdrop, optimizing computational efficiency. This holistic strategy culminated in insights that spotlight not just individual feature significance per timestep but also the collective interplay of features in steering route forecasts.

The culmination of our methodological approach yields multifaceted advantages. It doesn't merely visualize trajectories but also capacitates stakeholders to intuitively grasp and trace key model influencers. In contexts where the rationale behind a prediction rivals the prediction's importance, such an approach is invaluable. Moreover, this interpretability builds a formidable bridge of trust with users, reinforcing confidence in model outputs. The harmonious melding of attention mechanisms, LIME, and SHAP births a robust interpretability matrix, producing insights deeply rooted in the model while retaining the agility to span diverse models.

\begin{figure*}[h!]
\centering
\includegraphics[width=0.8\textwidth]{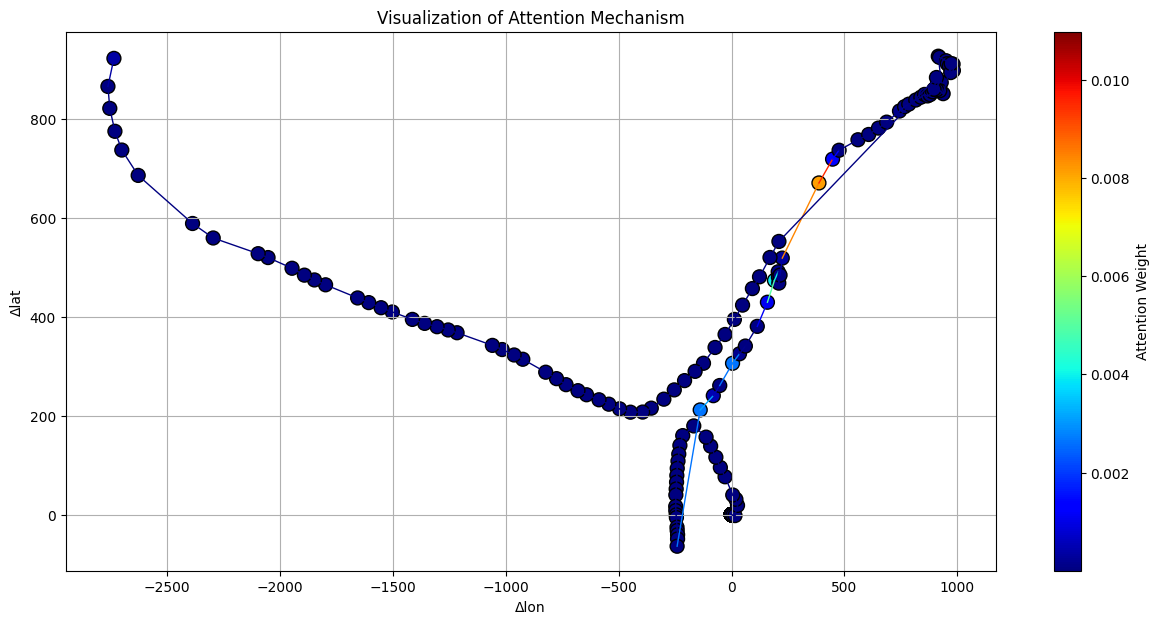}
\caption{ Attention mechanism-based highlighted trajectory. In this plot the historical trajectory is depicted and the most important parts of the historical route are also highlighted based on the attention layer weights.}
\label{fig:pfi}
\end{figure*}

\begin{figure*}[h!]
\centering
\includegraphics[width=0.8\textwidth]{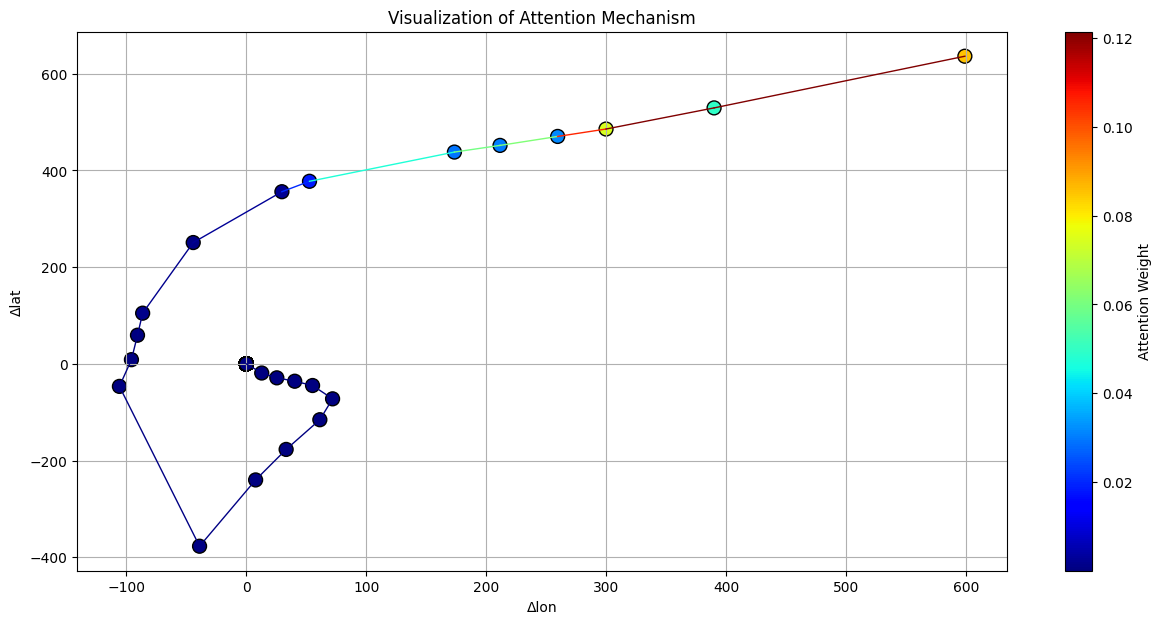}
\caption{LIME-based highlighted trajectory. In this plot the historical trajectory is depicted and the most important parts of the historical route are also highlighted based on the LIME weights. }
\label{fig:pfi}
\end{figure*}

In essence, our approach is a confluence of diverse interpretability techniques, each complementing the other, ensuring that stakeholders are not just informed but also enlightened about the inner dynamics of vessel route forecasting.

To achieve interpretability, we leverage the SHAP (SHapley Additive exPlanations) framework, which offers a unified measure of feature importance and inter-feature dependencies. By decomposing the prediction into a sum of feature contributions, SHAP values offer a clearer perspective on which features play pivotal roles and how they interact with one another to influence the model's decision.

Given the high-dimensional nature of our input data, we implement a tailored version of the Deep SHAP method. This method is mainly designed for deep learning models and can efficiently handle the intricacies of our bidirectional LSTM. A subset of the training data was chosen as the background dataset to accelerate the computation. Through this approach, not only do we determine the significance of each feature at every timestep but also discern how different features collectively influence route forecasting.

\begin{figure*}[h!]
\centering
\includegraphics[width=0.9\textwidth]{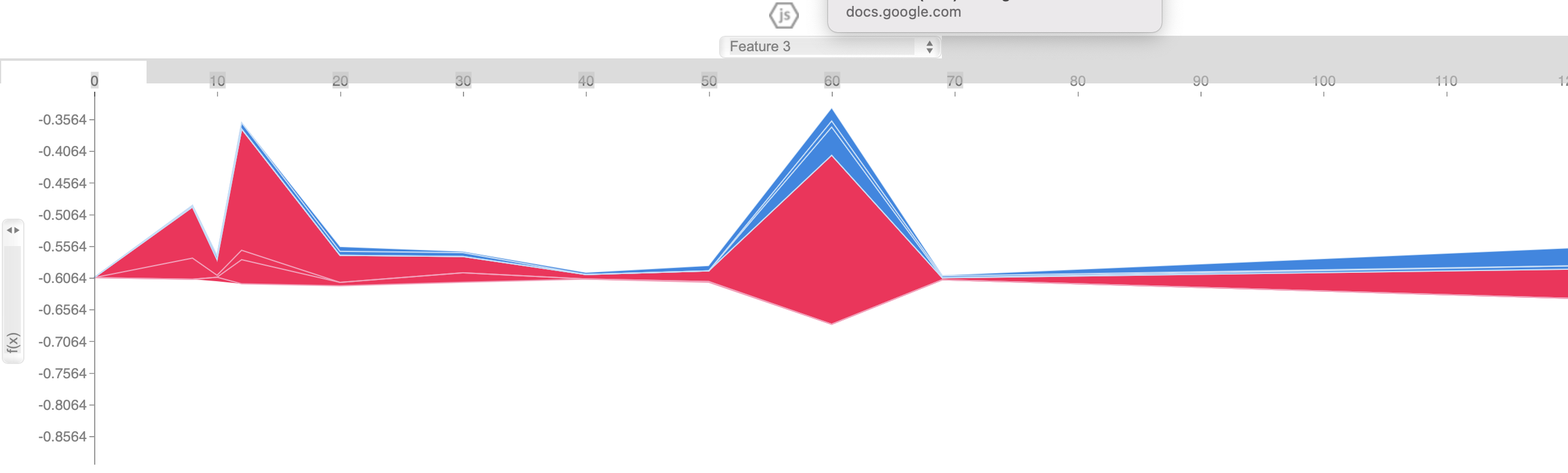}
\caption{SHAP values of in each time-step showing the importance of each feature at each time-step. }
\label{fig:pfi}
\end{figure*}

\begin{table}[h!]
\centering
\begin{tabularx}{\textwidth}{|X|X|X|}
\hline
\textbf{Aspect} & \textbf{General Results} & \textbf{Insights by Demographics} \\
\hline
Visualization for Accuracy & 
Strong preference for overlaying predicted vs. actual trajectories to assess model accuracy. & 
Universally appreciated across all groups.
Seen as a direct measure of model performance.
 \\
\hline
Visualization for Interpretability (XAI) & 
Positive feedback on highlighting important past positions of the vessel, providing insights into "why" a model made a prediction. & 
Technical roles found the LIME-like visualization of important past positions particularly insightful.
End-users felt empowered by such visualizations, finding them intuitive and helpful.
 \\
\hline
XAI Methods & 
PFI: Generally well-received
Saliency Maps: Mixed reception
SHAP: Positive traction
& 
Technical roles favored SHAP and PFI.
Non-technical roles leaned towards simpler explanations.
Academia displayed a balanced appreciation.
 \\
\hline
Trustworthiness & Desire for understanding 'why' behind predictions. Depth of understanding varied. & 
AI familiar respondents sought deep explanations.
End-users preferred intuitive explanations.
\\
\hline
\end{tabularx}
\caption{Summary of Results on XAI Usability in VRF}
\end{table}

\section{Results and Discussion}

In our endeavor to delve into the usability and effectiveness of XAI models within the domain of Vessel Route Forecasting (VRF), we administered a meticulously crafted survey to an eclectic mix of partners and participants, that it can be found \href{https://forms.gle/kS3AHLFwkXXXL4sWA}{here}. Drawing from real-world scenarios in major maritime corporations, we simulated a context where AI models transition from mere decision-support mechanisms to primary decision-makers, emphasizing their paramount need for transparency and trustworthiness. The survey pivoted on three primary axes: gauging participants' baseline familiarity with AI, ML, and DL; discerning their perception of key data interpretation and visualization techniques; and understanding their benchmarks for model trustworthiness, especially in terms of the model's time-dependent decision-making. By gathering feedback through this structured lens, our aim was to carve out a roadmap for the subsequent development and refinement of XAI models tailored to maritime applications.

Regarding the technical knowledge of the respondents,
the majority of them are familiar with AI (ML/DL) methods, and many also know about trajectory data.
However, not all are familiar or use AI for route forecasting, which could indicate that the application is niche, or the respondents' work doesn't always directly relate to this specific use-case.
There's a mix of professionals who use XAI (Explainable AI) methods and those who don't.

A pivotal focus was on visualization techniques, which are integral to comprehending and interpreting model outputs. Respondents displayed a strong preference for overlaying predicted versus actual trajectories. This method serves as a straightforward assessment of model accuracy and is universally appreciated across diverse professional backgrounds, marking it as a direct measure of model performance.

In terms of interpretability, another visualization method that gathered significant attention was the highlighting of important past positions of the vessel. This technique, based on LIME visualization, offers insights into the "why" behind model predictions. From the feedback, it was evident that technical roles, such as data scientists and ML engineers, found such visualizations particularly insightful. In contrast, end-users, who may not possess a deep technical background, felt empowered by these visualizations, marking them as intuitive in understanding AI-driven decisions.

On a deeper dive into specific XAI methods, PFI emerged as a generally well-received technique. Saliency Maps, however, received mixed feedback, pointing to potential areas of improvement or clarity. SHAP stood out with positive traction, especially among those familiar with AI and ML. This feedback underscores the balance between technical rigor and usability, a sentiment also evident in the academia's balanced appreciation of various XAI techniques.

To sum up, professionals from academic backgrounds, especially with roles like Data scientist - IT expert - ML engineer, tend to have a more technical understanding and often favor a combination of methods for interpretability.
End-users or those not familiar with AI and Data Science tend to have simpler preferences, like bar plots that show timestep importance or visualizations that highlight crucial parts of a vessel's trajectory.

\section{Conclusion}
In summing up the contributions of this paper, we tackle the intricate challenge of interpreting models based on vessel trajectory data—a domain teeming with unique complexities. While traditional XAI methods offer a wealth of insights, they sometimes stumble when faced with the nuances of spatio-temporal data. Our solution has been to form an integrative approach, combining the strengths of established XAI techniques: LIME, SHAP, Saliency maps, attention mechanisms, direct trajectory visualization, and Permutation Feature Importance.

Drawing your attention to our results, they clearly speak to the efficacy of our proposed methodology. Instead of employing these techniques in silos, we harmonize them to illuminate the intricate facets of trajectory data. This integrative approach ensures a comprehensive, yet nuanced, understanding of vessel routes. In essence, it's more than just demystifying model decisions—it's about painting a detailed, context-rich picture.

As we set our sights on the future, there's a dual focus. On one hand, we're eager to develop a system that translates XAI visuals into articulate textual explanations. On the other, we're inclined towards a more interactive visualization strategy, inviting users to delve deeper and offer feedback. It's not just about building models; it's about fostering a dialogue between human intuition and machine intelligence.

\textbf{Acknowledgements}

The research leading to the results presented in this paper has received funding from the Europeans Union’s funded Project MobiSpaces under grant agreement no 101070279.

\bibliographystyle{elsarticle-num} 
\bibliography{references}
\end{document}